\documentclass[conference]{IEEEtran}
\IEEEoverridecommandlockouts
\usepackage{amsfonts,amssymb}
\usepackage{setspace}
\usepackage{comment}
\usepackage{multirow}
\usepackage{booktabs}
\usepackage{amsmath,graphicx}
\usepackage[colorlinks,linkcolor=blue,anchorcolor=black,citecolor=blue]{hyperref} 

\graphicspath{{images/}}  

\begin{document}

% Title.
% ------
\title{MCTNet: A Multi-Scale CNN-Transformer Network for Change Detection in Optical Remote Sensing Images}
%
% Single address.
% ---------------
\author{\IEEEauthorblockN{Weiming Li, Lihui Xue, Xueqian Wang, and Gang Li*}
\IEEEauthorblockA{
}
}
%\ninept
%
\maketitle
\begin{abstract}
For the task of change detection (CD) in remote sensing images, deep convolution neural networks (CNNs)-based methods have recently aggregated transformer modules to improve the capability of global feature extraction. However, they suffer degraded CD performance on small changed areas due to the simple single-scale integration of deep CNNs and transformer modules. To address this issue, we propose a hybrid network based on multi-scale CNN-transformer structure, termed MCTNet, where the multi-scale global and local information is exploited to enhance the robustness of the CD performance on changed areas with different sizes. Especially, we design the ConvTrans block to adaptively aggregate global features from transformer modules and local features from CNN layers, which provides abundant global-local features with different scales. Experimental results demonstrate that our MCTNet achieves better detection performance than existing state-of-the-art CD methods.
\end{abstract}
\section{Introduction}
\label{sec:intro}

Change detection (CD) in optical remote sensing images has received extensive attention due to its key role in the context of earth observation and environmental monitoring. It aims to detect changes in two or more remote sensing images observing the same geographical areas captured at different periods. With the recent development of spaceborne/airborne optical imaging technologies, CD using optical remote sensing images has become one of the most significant tasks in disaster assessment~\cite{jiang2021semisupervised}, urban change investigation~\cite{jiang2019building}, etc.

In recent years, with the development of deep learning (DL) technologies and the increase of optical remote sensing data, many DL-based CD methods have been proposed and outperformed the traditional CD methods based on handcrafted features owing to their powerful and automatic feature extraction capabilities. Convolutional neural networks (CNNs) have been widely used in existing DL-based CD methods to extract features of remote sensing images due to the strong local modeling ability of convolution operations. Daudt \emph{et al.}~\cite{daudt2018fully} developed three fully convolutional CD networks, where U-Net~\cite{ronneberger2015u} was integrated with the Siamese encoder for the first time to fuse the bi-temporal features and obtain changed areas. Peng \emph{et al.}~\cite{peng2019end} and Fang \emph{et al.}~\cite{fang2021snunet} improved Daudt's work by extending sparse skip connections to dense skip connections, which helps the networks reduce loss of localization information and achieve better boundary prediction. However, due to the inherent locality of convolution operations and limited reception fields, the aforementioned methods~\cite{daudt2018fully,peng2019end,fang2021snunet} based on pure convolution structures are difficult to understand change patterns from the global perspective, which makes them hardly distinguish between real changes and pseudo changes under the interference of different imaging conditions (e.g., illumination variation and atmospheric environments) in bi-temporal images.

Transformer models~\cite{vaswani2017attention,dosovitskiy2020image,wang2021pyramid,wu2021cvt} have been applied in multiple computer vision tasks because of their great advantage to capture global information in images. Note that, despite the transformer's superiority to model global context, the self-attention mechanism in transformer often hardly captures fine-grained local details~\cite{yuan2021tokens}. Inspired by the respective strengths of CNN/transformer in local/global feature representations, Chen \emph{et al.}~\cite{chen2021remote} proposed the bi-temporal image transformer (BIT) method, where transformer is introduced into the CD task to extract global context from the high-level features obtained by deep CNN. However, BIT~\cite{chen2021remote} only adopts the simple single-scale integration of deep CNNs and transformer, where transformer is performed after deep CNNs and its extracted global features cannot contain the small targets missed after the preceding deep CNN structure. Therefore, BIT cannot be robust to CD tasks regarding small changed areas.

\begin{figure*}[!h]
\centering
\includegraphics[width=0.95\textwidth]{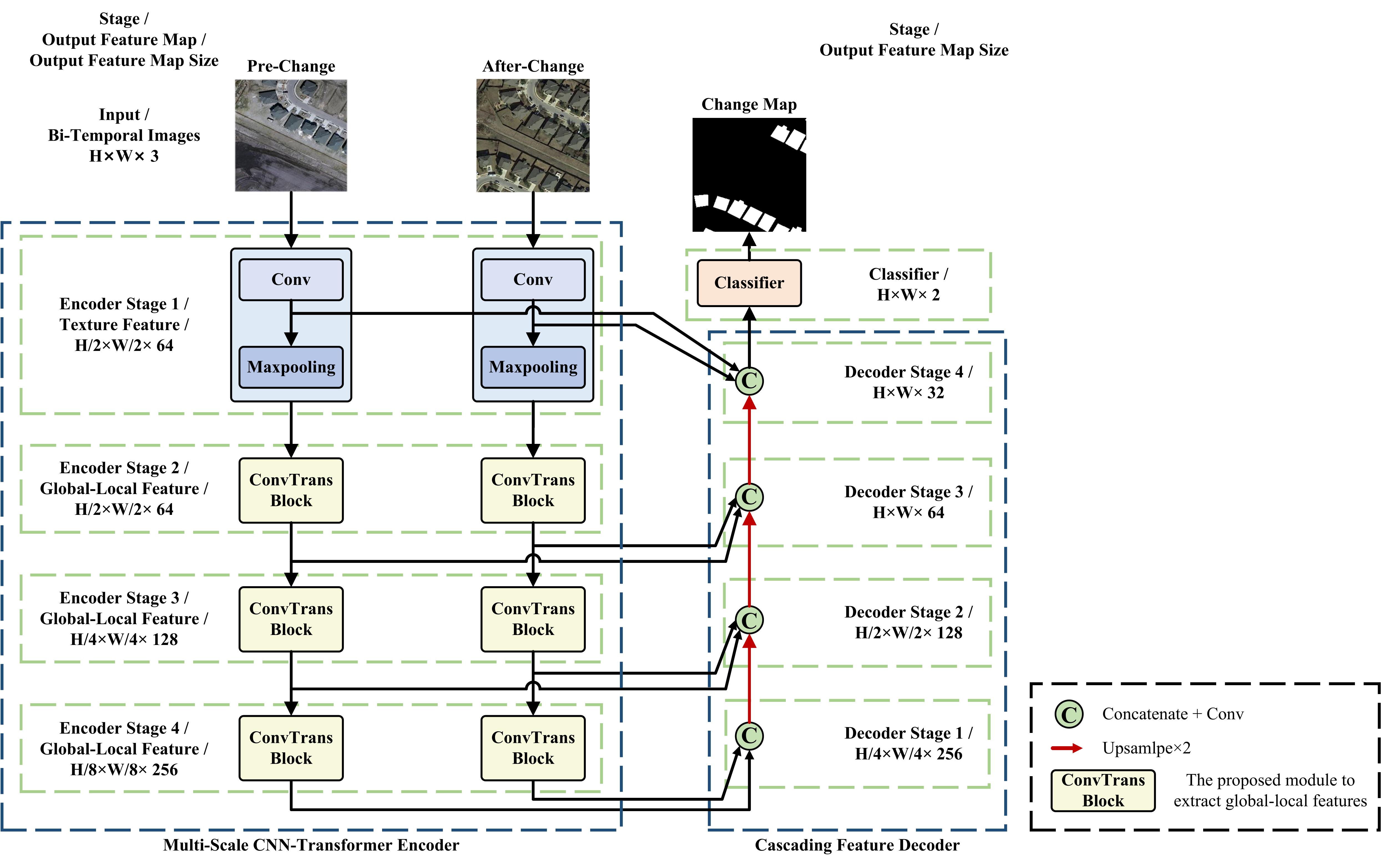}
\caption{Architecture of the proposed MCTNet} \label{fig1}
\end{figure*}

In this paper, a new CD method termed MCTNet is proposed to integrate CNN and transformer in the multi-scale fashion while simultaneously exploiting multi-scale global-local information in bi-temporal optical remote sensing images. Especially, a dual-stream feature extraction unit, namely ConvTrans block, is designed to capture more comprehensive image features from the global-local perspective. In our MCTNet, we first construct a Siamese encoder via multiple ConvTrans blocks to capture the multi-scale global-local features from remote sensing images. Then, the multi-scale global-local features are fused layer by layer in a decoder with skip connections~\cite{ronneberger2015u} to enhance the saliency of the changed areas. Based on these, our method detects the changed areas with different sizes, notably the small changed areas more robustly, leading to a significant boost on CD performance. Experiments based on real data demonstrate the advantages of our newly proposed method. 

\section{Proposed MCTNet Method}
\label{sec:method}
\subsection{Overview architecture}
The overall architecture of the proposed MCTNet is illustrated in Fig.~\ref{fig1}. MCTNet is a standard encoder-decoder architecture including two parts, i.e., an encoder with the Siamese structure for multi-scale global-local feature extraction and a decoder with the skip connections~\cite{ronneberger2015u} for multi-scale global-local feature fusion.
\begin{figure*}[!h]
\centering
\includegraphics[width=0.60\textwidth]{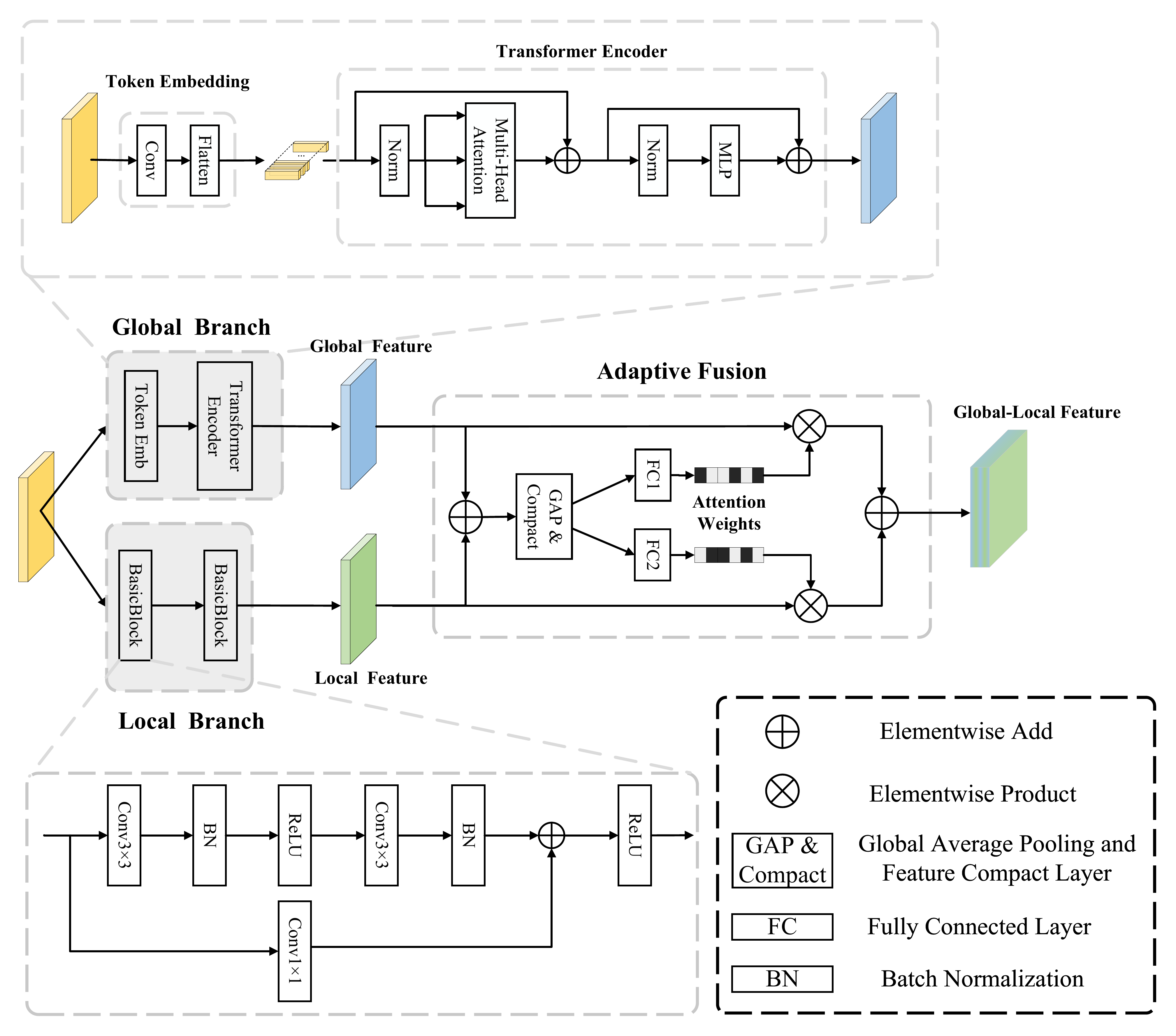}
\caption{Structure of the proposed ConvTrans block} \label{fig2}
\end{figure*}

Suppose that ${I^{{t_1}}} \in {\mathbb{R}^{H \times W \times 3}}$ and ${I^{{t_2}}} \in {\mathbb{R}^{H \times W \times 3}}$ are the bi-temporal remote sensing images captured in the same region at different times ${t_1}$ and ${t_2}$. The goal of CD is to generate a pixel-wise binary change map that points out the changed regions between ${I^{{t_1}}}$ and ${I^{{t_2}}}$. As seen in Fig.~\ref{fig1}, first, the bi-temporal images ${I^{{t_1}}}$ and ${I^{{t_2}}}$ are input into a Siamese encoder, which contains two parallel subnetworks with identical structures and shared weights. Each subnetwork consists of four stages: one convolution block and three ConvTrans blocks. In the first encoder stage, a convolution operation is utilized to extract the low-level features (e.g., edge and texture information) from remote sensing images. Then, the proposed ConvTrans blocks utilized in encoder stages 2-4 (see Fig.~\ref{fig1}) effectively extract the multi-scale global-local features. Next, the decoder with the skip connections~\cite{ronneberger2015u} is used to fuse the multi-scale global-local features from the Siamese encoder layer by layer to progressively enhance the saliency of the changed areas. Finally, a classifier with two 1$\times$1 convolution layers is used to produce the change map. In addition, during the process of training, the weighted cross-entropy loss is employed to optimize the parameters of our MCTNet.  

\subsection{ConvTrans block}
The ConvTrans block is designed as the feature extraction unit of our MCTNet to effectively extract the global-local features in remote sensing images. The structure of the ConvTrans block is illustrated in Fig.~\ref{fig2}, the ConvTrans block consists of three parts: local branch based on convolution operations to extract the local features, global branch based on the transformer~\cite{wu2021cvt} to summarize the global features, and an adaptive fusion module~\cite{li2019selective} to aggregate the global features and the local features into the global-local features. 

\subsubsection{Local branch}
Two sequential BasicBlocks in~\cite{he2016deep} make up the local branch to capture the local features $F_i^{local}$ in the $i$-th encoder stage, each of them containing two 3$\times$3 convolution operations. The process of a BasicBlock $H( \cdot )$ is formulated as:
\begin{equation}
H(x) = \sigma(BN({W_2}(\sigma (BN({W_1}x)))) + {W^I}x)
\end{equation}
where ${W_1}$ and ${W_2}$ represent the weights of two 3$\times$3 convolution operations, respectively. ${W^I}$ denotes the weights of a 1$\times$1 convolution operation. Note that when there is no down-sampling requirement in the current BasicBlock, the weights in ${W^I}$ are equal to 1 (i.e., an identical mapping for $x$). $\sigma ( \cdot )$ is the rectified linear unit ($ReLU$)~\cite{krizhevsky2012imagenet} and $BN( \cdot )$ is the batch normalization layer~\cite{ioffe2015batch}.

\subsubsection{Global branch}
The global branch consists of a token embedding module and a transformer encoder referred from~\cite{wu2021cvt}.  Taking the output feature maps ${F_{i - 1}} \in {\mathbb{R}^{{H_{i - 1}} \times {W_{i - 1}} \times {C_{i - 1}}}}$ from the ${i-1}$-th encoder stage as input, the token embedding module in the $i$-th encoder stage uses a convolution operation $h_i( \cdot )$ to convert ${F_{i - 1}}$ into a token map with channel size ${C_i}$. Then, the obtained token map $h_i({F_{i - 1}}) \in \mathbb{R}^{^{{H_i} \times {W_i} \times {C_i}}}$ is flattened into a token sequence ${T_i}$ with 1D shape (where $i$ ranges from 2 to 4). 

After obtaining the tokens ${T_i}$, we model the global context among the tokens via the transformer encoder. In the transformer encoder, the obtained tokens ${T_i}$ are first normalized by layer normalization~\cite{ba2016layer}. Then, the linear projection layers are used to generate three vectors, termed query $Q$, key $K$, and value $V$. Subsequently, the multi-head self-attention (MSA)~\cite{vaswani2017attention} separates $Q$, $K$, and $V$ into multiple independent heads and calculates the attention coefficients of each head in parallel. The outputs of all heads are concatenated and then linearly projected to generate the enhanced tokens ${\hat T_i}$  in the $i$-th encoder stage:
\begin{equation}
\begin{split}
& Attention(Q,K,V) = softmax(\frac{{Q{K^T}}}{{\sqrt {{d_{head}}} }})V \\
& head{_j} = Attention({Q_j,K_j,V_j}) \\
& {{\hat T}_i} = Concat(head{_1}, \ldots ,head{_n}){W^O} \\
\end{split}
\end{equation}
where ${W^O}$ represents the weights of a linear projection layer. $n$ is the number of heads in MSA and $j$ ranges from 1 to $n$ . ${d_{head}}$ represents the channels of each head and equals to ${C_i}/{n}$. Finally, the multi-layer perceptron (MLP) module is leveraged to further enhance the representation of the tokens. Assuming the input of MLP is $\bar{T}_i$, the process of the MLP module is formulated as:
\begin{equation}
\begin{split}
& g(x) = Flatten(Reshape2D(x){W^D}) \\
& MLP({{\bar{T}_i}}) = GELU(g({\bar{T}_i}{W_3})){W_4} \\
\end{split}
\end{equation}
where ${W_3}$, ${W_4}$ are the linear projection matrices to expand and compress the channel dimensions of the tokens, respectively. ${W^D}$ is the weight matrix of a depth-wise convolution operation \cite{howard2017mobilenets} for positional information encoding \cite{valanarasu2022unext}. $GELU( \cdot )$~\cite{hendrycks2016gaussian} denotes the activation function of gaussian error linear units. Finally, the output tokens of MLP with residual connection are reshaped into a 2D feature map $F_i^{global}$ containing rich global context.
\begin{figure*}[t]
\centering
\includegraphics[width=1\textwidth]{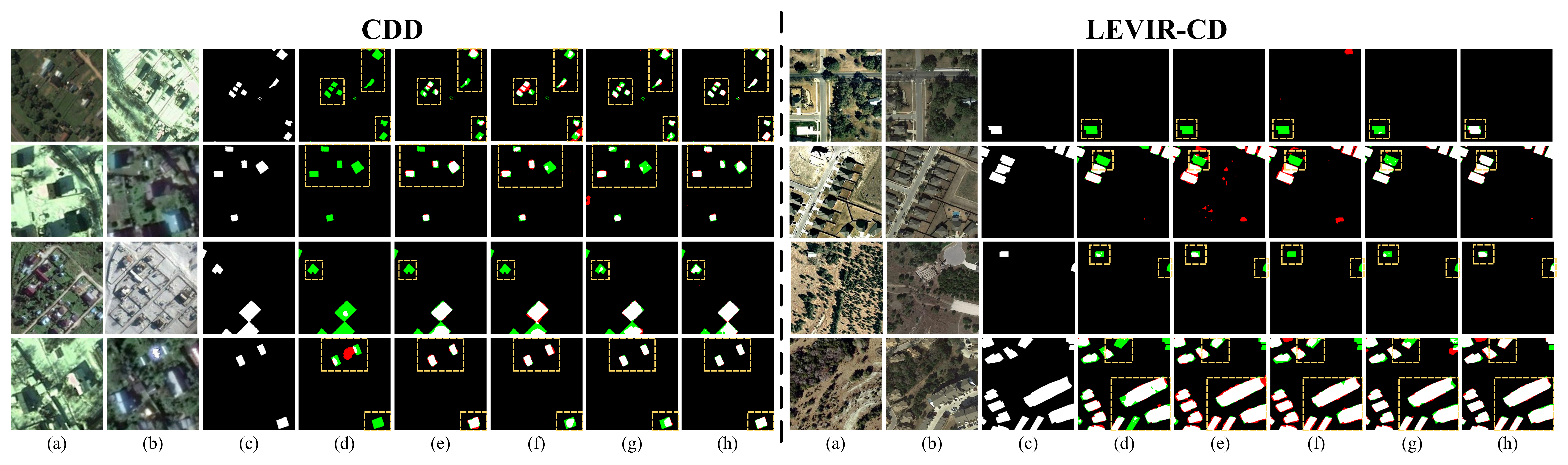}
\caption{Visualized results on CDD and LEVIR-CD. (a) Image shot before event, (b) Image shot after event, (c) Ground truth, (d) FC-Siam-Di~\cite{daudt2018fully}, (e) DASNet~\cite{chen2020dasnet}, (f) STANet~\cite{chen2020spatial}, (g) BIT~\cite{chen2021remote}, and (h) our MCTNet. White, black, red, and green areas indicate true positive, true negative, false positive, and false negative predictions, respectively.} 
\label{fig3}
\end{figure*}
\subsubsection{Adaptive fusion}
After obtaining the local features $F_i^{local}$ and the global features $F_i^{global}$, an important problem is how to effectively fuse the features from the global and local branches carrying the information with different sizes of reception fields. Here, we introduce an automatic approach in~\cite{li2019selective} to aggregate the features from the local branch and the global branch in an adaptive manner. The adaptive fusion module contains three steps: feature integration, attention calculation, and feature selection. Firstly, the adaptive fusion module integrates the features from the local branch and the global branch via element-wise summation, and then the integrated features are embedded into the channel-wise space via a global average pooling (GAP) operation and a feature compact layer. Next, two fully connected (FC) layers and a softmax operation are used to produce two adaptive attention weights for the local branch and the global branch with the help of channel-wise embedding features, respectively. Finally, the global-local feature ${F_i}$ is formulated by using the attention weights to fuse the global and local feature representations.

%%%%%%%%%%%%%%%%%%%%%%%%%%%%%%%%%%%%%%%%%%%%%%%%%%%%%%%%%%%%%%%%%%%%%%%%%%%%%%%%
\section{Experiment Results}
\label{sec:experiment}
\subsection{Dataset}
To assess the performance of our MCTNet, in-depth comparative experiments are conducted on two typical datasets named CDD~\cite{lebedev2018change} and LEVIR-CD~\cite{chen2020spatial}. The CDD dataset comprises 11 pairs of optical season-varying remote sensing images that cover various changes, such as trees, buildings, roads, etc. LEVIR-CD contains 637 pairs of high-resolution remote sensing images which mainly focus on building changes. All images are cropped into small patches with the size of 256$\times$256. In our experiment, 10000/2998/3000 and 7120/1024/2048 pairs of image patches in CDD~\cite{lebedev2018change} and LEVIR-CD~\cite{chen2020spatial}, respectively, are to train/validate/test the existing CD algorithms, i.e., FC-Siam-Di~\cite{daudt2018fully}, DASNet~\cite{chen2020dasnet}, STANet~\cite{chen2020spatial}, BIT~\cite{chen2021remote}, and our proposed MCTNet. The above SOTA methods are re-implemented via their available codes.

\subsection{Implementation details}
We implement our MCTNet by Pytorch and train it on a workstation with two NVIDIA RTX TITAN GPUs. The stochastic gradient descent (SGD) method with momentum is utilized to optimize our model, and the momentum and weight decay are set to 0.99 and 0.001, respectively. The initial learning rate is set to 0.001, and it is subsequently reduced to 10\% of its original value after one-third of the total epochs.

\subsection{Performance comparisons}
We utilize four indicators also used in ~\cite{daudt2018fully,chen2021remote,chen2020dasnet,chen2020spatial} to evaluate the performance of our MCTNet: precision, recall, overall accuracy (OA), and F1-score. Note that larger values of these four metrics, especially F1-score and OA, indicate better CD performance. The quantitative comparisons among different CD methods are presented in Table~\ref{tab1} and Table~\ref{tab2}. From the tables, we can observe that our MCTNet achieves the higher value on the evaluation indicators. Compared with DASNet~\cite{chen2020dasnet} and STANet~\cite{chen2020spatial} based on the specific attention modules, our method improves F1-score by 3.2\%/4.7\% respectively on CDD and 5.3\%/4.5\% respectively on LEVIR-CD. Compared with BIT~\cite{chen2021remote} which considers global context via the transformer module on the single-scale features from deep CNNs, our method boosts F1-score by 1.9\% and 1.5\% on CDD and LEVIR-CD, respectively. The main reason for this noticeable improvement is that we fully exploit global and local information in multi-scale aspects, pushing our method to produce more accurate and complete CD results on the changed areas with different sizes.
\begin{table}
\centering
\caption{Quantitative Comparisons on CDD. The Average Performance of Three Experiments Is Reported for Different Methods.}
\label{tab1}
\resizebox{0.5\textwidth}{!}{
\begin{tabular}{ccccccccccccccccc} %需要5列
\toprule
Method & \multicolumn{3}{c}{} & Precision & \multicolumn{3}{c}{} & Recall & \multicolumn{3}{c}{} & F1-score & \multicolumn{3}{c}{} & OA	\\
\midrule
FC-Siam-Di ~\cite{daudt2018fully} & \multicolumn{3}{c}{} & 87.51\%  & \multicolumn{3}{c}{} & 59.82\% & \multicolumn{3}{c}{} & 71.06\% & \multicolumn{3}{c}{} & 94.14\% \\
DASNet ~\cite{chen2020dasnet} & \multicolumn{3}{c}{} & 93.62\%  & \multicolumn{3}{c}{} & 92.14\% & \multicolumn{3}{c}{} & 92.87\% & \multicolumn{3}{c}{} & 98.30\% \\
STANet ~\cite{chen2020spatial} & \multicolumn{3}{c}{} & 88.78\%  & \multicolumn{3}{c}{} & 93.98\% & \multicolumn{3}{c}{} & 91.31\% & \multicolumn{3}{c}{} & 97.85\% \\
BIT ~\cite{chen2021remote} & \multicolumn{3}{c}{} & 95.89\%  & \multicolumn{3}{c}{} & 92.48\% & \multicolumn{3}{c}{} & 94.15\% & \multicolumn{3}{c}{} & 98.62\% \\
\textbf{MCTNet} & \multicolumn{3}{c}{} & \textbf{97.59\%}  & \multicolumn{3}{c}{} & \textbf{94.63\%} & \multicolumn{3}{c}{} & \textbf{96.09\%} & \multicolumn{3}{c}{} & \textbf{99.07\%} \\
\bottomrule
\end{tabular}
}
\end{table}
\begin{table}
\centering
\caption{Quantitative Comparisons on LEVIR-CD. The Average Performance of Three Experiments Is Reported for Different Methods.}
\label{tab2}
\resizebox{0.5\textwidth}{!}{
\begin{tabular}{ccccccccccccccccc} %需要5列
\toprule
Method & \multicolumn{3}{c}{} & Precision & \multicolumn{3}{c}{} & Recall & \multicolumn{3}{c}{} & F1-score & \multicolumn{3}{c}{} & OA	\\
\midrule
FC-Siam-Di ~\cite{daudt2018fully} & \multicolumn{3}{c}{} & 89.36\%  & \multicolumn{3}{c}{} & 84.07\% & \multicolumn{3}{c}{} & 86.63\% & \multicolumn{3}{c}{} & 98.67\% \\
DASNet ~\cite{chen2020dasnet} & \multicolumn{3}{c}{} & 81.68\%  & \multicolumn{3}{c}{} & 89.91\% & \multicolumn{3}{c}{} & 85.60\% & \multicolumn{3}{c}{} & 98.45\% \\
STANet ~\cite{chen2020spatial} & \multicolumn{3}{c}{} & 82.65\%  & \multicolumn{3}{c}{} & \textbf{90.34\%} & \multicolumn{3}{c}{} & 86.32\% & \multicolumn{3}{c}{} & 98.54\% \\
BIT ~\cite{chen2021remote} & \multicolumn{3}{c}{} & 91.35\%  & \multicolumn{3}{c}{} & 87.43\% & \multicolumn{3}{c}{} & 89.35\% & \multicolumn{3}{c}{} & 98.93\% \\
\textbf{MCTNet} & \multicolumn{3}{c}{} & \textbf{91.57\%}  & \multicolumn{3}{c}{} & 90.26\% & \multicolumn{3}{c}{} & \textbf{90.91\%} & \multicolumn{3}{c}{} & \textbf{99.08\%} \\
\bottomrule
\end{tabular}
}
\end{table}

In Fig.~\ref{fig3}, we display the visualized detection results of different methods. It can be observed that FC-Siam-Di~\cite{daudt2018fully} is critically affected by light intensity and season alternation, leading to many omissions in the results. STANet~\cite{chen2020spatial} and DASNet~\cite{chen2020dasnet} alleviate this problem via the specific attention modules. However, the spatial information deteriorates in the attention modules so there are a number of boundary false alarms in the detection results of STANet~\cite{chen2020spatial} and DASNet~\cite{chen2020dasnet}. BIT~\cite{chen2021remote} captures global context in the coarse-grained high-level features from the deep layer of CNN, achieving an improved detection under the varying imaging conditions, but it fails to recognize the small changed areas due to the neglection of multi-scale information. In contrast, our method detects the changes with different sizes, especially the small changed areas (see the first, second, and third rows of Fig.~\ref{fig3}) more completely and accurately, which attributes to our effective design for extracting and fusing the multi-scale global-local features.

\section{Conclusion}
\label{sec:conclusion}
In this article, we propose MCTNet, a network with multiscale CNN-transformer architecture for CD in optical remote sensing images. In MCTNet, we mainly design a dual-stream ConvTrans block, which can effectively extract global and local features from remote sensing images and then adaptively aggregate them to formulate the global-local features. Compared to the existing CD methods, our MCTNet better exploits global-local information on multiple scales, enhancing robustness of the CD performance on changes with different sizes, especially for small changed areas. Experiment results verify the effectiveness of our MCTNet. Our future work includes the investigation and modification of MCTNet in the heterogenous CD scenarios with both optical and synthetic aperture radar images. 
%%%%%%%%%%%%%%%%%%%%%%%%%%%%%%%%%%%%%%%%%%%%%%%%%%%%%%%%%%%%%%%%%%%%%%%%%%%%%%%%
\bibliographystyle{IEEEbib}
\bibliography{strings,refs}

\end{document}